\documentclass[letterpaper, 10 pt, conference]{ieeeconf}  %

\IEEEoverridecommandlockouts                              %

\usepackage{graphics} %
\usepackage{epsfig} %
\usepackage{amsmath} %
\usepackage{amssymb}  %

\usepackage{booktabs}
\usepackage{hyperref}

\usepackage{cite}
\usepackage{array}
\usepackage{multirow}
\usepackage[dvipsnames]{xcolor}
\usepackage{colortbl}
\usepackage{tikz}

\newcommand{\xmark}{%
\tikz[scale=0.12] {
    \draw[line width=0.6,line cap=round] (0,0) to [bend left=6] (1,1);
    \draw[line width=0.6,line cap=round] (0.2,0.95) to [bend right=3] (0.8,0.05);
}}
\newcommand{\cmark}{%
\tikz[scale=0.2] {
    \draw[line width=0.6,line cap=round] (0.25,0) to [bend left=10] (1,1);
    \draw[line width=0.7,line cap=round] (0,0.35) to [bend right=1] (0.23,0);
}}

\definecolor{mygray}{rgb}{0.6, 0.6, 0.6}

\usepackage[capitalise]{cleveref}
\crefname{section}{Sec.}{Secs.}

\usepackage{siunitx}

\usepackage{tablefootnote}

\usepackage{etoolbox}
\makeatletter
\patchcmd{\@makecaption}
  {\scshape}
  {}
  {}
  {}
\makeatother

\title{\LARGE \bf
A comparison of visual representations for real-world reinforcement learning in the context of vacuum gripping
}

\author{
    Nico Sutter, Valentin N. Hartmann, Stelian Coros%
    \thanks{All authors are with the Computational Robotics Lab, ETH Zurich, CH.}%
}

\newcolumntype{s}{>{\scriptsize}l}

\begin{document}

\maketitle
\thispagestyle{empty}
\pagestyle{empty}

\begin{abstract}
\label{sec:abstract}
When manipulating objects in the real world, we need reactive feedback policies that take into account sensor information to inform decisions.
This study aims to determine how different encoders can be used in a reinforcement learning (RL) framework to interpret the spatial environment in the local surroundings of a robot arm.
Our investigation focuses on comparing real-world vision with 3D scene inputs, exploring new architectures in the process.
We built on the SERL framework, providing us with a sample efficient and stable RL foundation we could build upon, while keeping training times minimal.
The results of this study indicate that spatial information helps to significantly outperform the visual counterpart, tested on a box picking task with a vacuum gripper.
The code and videos of the evaluations are available at \url{https://github.com/nisutte/voxel-serl}.

\end{abstract}

\section{Introduction}
\label{sec:introduction}

Understanding and interpreting complex environments and spatial relations is vital for the effective deployment of robotic systems across a variety of tasks. 
This work examines how different real-world vision and 3D scene inputs can be integrated into reinforcement learning policies to enhance the robot's ability to perceive the spatial environment, enabling more accurate interaction with its surroundings. 
We evaluate these approaches on the deceptively challenging task of using a vacuum gripper to pick up parcels of various sizes, shapes and weights.
Vacuum grippers are a widely used tool in industrial automation for object manipulation, as they can handle a diverse range of objects due to their ability to create a secure grip over large surface areas, as long as the gripping position is chosen properly.
They can however fail just as easily and fail to establish a secure grasp in case the object is deformed or the surface is not smooth enough, and the suction cup can not thus form a seal.
This can be partially mitigated by using multiple suction cups on an end-effector in order to increase the chances of successfully engaging the gripper.

In this work, we focus on the final phase of the manipulation problem using only a single vacuum gripper.
We assume that we have access to a rough estimate of the box's location and orientation, and concentrate on 'last-inch manipulation', and the intricacies of the gripping process, rather than the overall movement to the object of interest.
Given the potential for inaccurate estimates or situations where the boxes are damaged or partially occluded, having an adaptive local feedback policy is essential.

Simulating the dynamics of deformable boxes accurately poses significant challenges due to the complex interactions between the deformable vacuum gripper and the deformable objects. 
These interactions are influenced by factors such as suction cup elasticity, surface texture, and object geometry, making accurate simulation difficult.

These obstacles underscore the need for real-world training of reinforcement learning (RL) policies to ensure robust performance, as simulated environments often fail to capture the intricacies of physical interactions, leading to significant sim-to-real gaps. 
Our work demonstrates the effectiveness of these policies in managing the complexities associated with vacuum gripper tasks, offering insights into their broader applicability in robotics.

Our work builds on the SERL (Sample-Efficient robotic Reinforcement Learning) framework \cite{luo2024serl}, a software suite for robotic RL with sample-efficient off-policy algorithms, maximizing learning efficiency in an environment where data collection is costly and time-consuming. 

\begin{figure}[t!]
  \centering
  \includegraphics[width=0.95\linewidth]{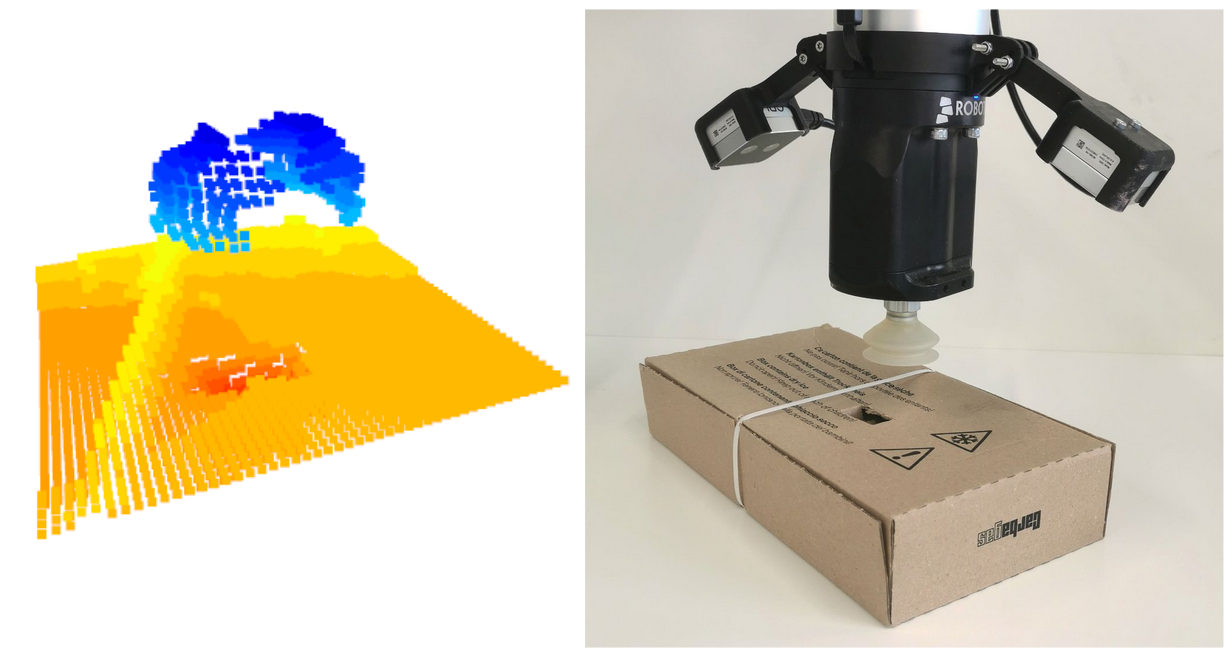}
  \caption{\textbf{(left)} Voxel grid representation of the suction gripper and the box that is used as \emph{spatial} observation for the learned policy. \textbf{(right)} Robot arm with vacuum gripper grasping a box.}
  \label{fig:intro}
\end{figure}

In this work, we compare two methods to incorporate spatial information, depth images and a voxel grid representation (\cref{fig:intro}), to using only visual information.
Using an encoder based on 3D convolutions following \cite{maturana2015voxnet}, the agent can capture the 3D structure of the environment more effectively.
We show that this approach not only improves the agent's ability to generalize across different objects, but also enhances its performance in complex and unstructured environments.

For the visual component of our RL system, we use pre-trained image encoders to extract meaningful features from visual inputs. 
These encoders provide the agent with the ability to perceive the environment, enabling it to make informed decisions during the grasping process. 
However, relying solely on 2D visual data can limit the agent's ability to generalize across different objects and scenarios \cite{ling2023efficacy}.

Through the integration of these spatial perception techniques with SERL, our work aims to develop a robust RL framework capable of efficiently training real-world robotic systems for reliable object grasping using vacuum grippers.

Summarizing, our main contributions are (1) a comparison of different visual and spatial observations in our problem setting, and  (2) testing the spatial encoder in a real world application.

\section{Related Work}
\label{sec:related work}
We will here discuss work on incorporating visual and spatial observations in learning. 
For a more comprehensive overview on real world RL for robotics, we refer to \cite{zhu2020ingredients, Kormushev2013rl}.

Incorporating \textbf{vision in robot learning } led to significant progress in robotic navigation \cite{zhu2021deep, desouza2002vision}, manipulation \cite{nair2022relw_visrl, kalashnikov2018relw_visrl} and grasping \cite{martinez2019vision, du2021vision,cheng2022vision}. 
In RL, a common approach to incorporate vision observations is the use of a pre-trained vision backbone \cite{shah2021rrl, spector2021insertionnet, luo2024serl, xiao2022masked}.
A limitation of this approach is a possible data-mismatch between robot-environments and human-centric environments.
The work in \cite{nair2022relw_visrl} then introduces a universal visual representation focused specifically on robot manipulation. It has been demonstrated to work well for imitation learning, but has not yet been tested in RL settings.

It is also possible to train the full network from scratch, as e.g. in \cite{kalashnikov2018relw_visrl} where a a deep network is trained for closed loop control of robotic grasping.
However, this method requires large datasets, long training times, and it is not clear how well this method generalizes to other settings with e.g., different lighting conditions.
It has been shown that data augmentation improves sample efficiency of visual agents, \cite{yarats2021imageDRQ, laskin2020relw_visrl, xu2023relw_visrl}, possibly helping alleviate this problem.

\textbf{Spatial representations in robot learning} have been used for localization, segmentation, and pose estimation \cite{du2021vision} before, mainly relying on point cloud representations.
Point cloud representations have also been widely used in vision-based robotic grasping, especially in grasp proposal estimation \cite{ten2017grasp, alliegro2022relw_end, sundermeyer2021relw_contact}.
However, there are instances where a stable grasp pose is proposed, but it is not achievable by planning.
To achieve efficient grasping, various reinforcement learning policies with point cloud inputs have been demonstrated in simulation \cite{chen2022system, huang2021generalization, wu2023learning}, which allows the agent to make informed, reactive decisions.

The work that is most similar to ours is \cite{ling2023efficacy}, which compares 2D image observations with spatial observations in RL, finding that 3D point cloud encoders can significantly outperform 2D observations.
However, the 3D point clouds that were used as input in \cite{ling2023efficacy} represent the whole scene, which is assumed to be known, compared to our work, where we only observe parts of the environment with a sensor.
Further, these comparisons were only made in simulation and not shown on a real robot.

\section{Background}
\label{sec:algo selection}
We formulate the general RL task as a partially observable Markov decision process (POMDP) \cite{bellman1957markovian, kaelbling1998planning}. 
A POMDP can be described as a tuple $(\mathcal{O, A}, p, r, \gamma)$, where $\mathcal{O}$ is the observation space, i.e., the robot states as well as higher-dimensional representations, $\mathcal{A}$ is the action space, $p$ is the unknown and potentially stochastic transition probability that depends on the system dynamics, $r: \mathcal{O} \times \mathcal{A} \rightarrow \mathbb{R}$ is the reward function that maps the current observation and action to reward $r_t = r(o_{t}, a_t)$ and $\gamma$ is the discount factor.
Our goal is to find an optimal policy $\pi(a_t | o_t)$, that maximizes the expected accumulated return given by $\mathbb{E}_{\pi} \sum^{\infty}_{t=0} \gamma^t r_t$.

\subsection{Behavior Tree}
\label{sec:BT}
\begin{figure}[t]
  \centering
  \includegraphics[width=0.8\linewidth]{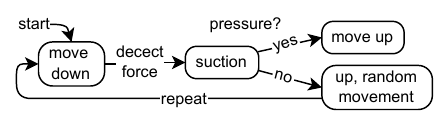}
  \caption{Illustration of the Behavior Tree for grasping a box.}
  \label{graphics:behavior_tree}
\end{figure}

As baseline for the task, we use a simple behavior tree which is depicted in \cref{graphics:behavior_tree}.  
It descends whenever possible and enables the suction gripper upon detecting a force in the negative Z-axis. 
If the attempt is successful, the agent moves up and is finished if the grasp is still active, else it ascends, changes its position randomly in the XY-plane and reattempts.
\subsection{Behavioral cloning}
\label{sec:BC}
For imitation learning from human demonstrations, behavioral cloning (BC) \cite{pomerleau1988alvinn} is one of the most well known and widely used algorithms, mainly due to its simplicity. BC is implemented using the following objective:
\begin{equation}
\hat{\mathbf{\theta}} = \arg\max_{\theta} \mathbb{E}_{(s,a) \sim \tau_E} \left[ \log (\pi_{\theta}(a|s)) \right]
\end{equation}
where $s$ is the continuous state, $a$ is the continuous action, $\tau_E$ represents the state action trajectories of the demonstrations and $\pi_\theta$ is the current policy. 

\subsection{Soft Actor-Critic}
\label{sec:SAC}
SAC \cite{haarnoja2018softAC} is an RL algorithm for continuous control tasks that optimizes a stochastic policy $\pi_\theta$ by maximizing both policy entropy \cite{ziebart2008maximum} and expected reward via the state-action value function $Q_\theta$, encouraging exploration.  
Since SAC cannot effectively handle high-dimensional observation spaces like images or point-cloud data, we use this state-of-the-art off-policy algorithm as a baseline for comparison with more advanced policies.

\subsection{Reinforcement Learning with Prior Data (RLPD)}
To make real-world RL efficient, SERL builds upon RLPD \cite{ball2023efficient}, a recently proposed algorithm which is based on SAC.
Key modifications to SAC include: (1) high update-to-data ratio during training, (2) symmetric sampling of prior and on-policy data, and (3) layer-norm regularization during training.
While this method can train from scratch, it is often useful to bootstrap learning using prior data (e.g. demonstrations).

\subsection{Data-regularized Q}
\label{sec:DRQ}
DRQ \cite{yarats2021imageDRQ} is a simple data augmentation technique that can be applied to any standard model-free RL algorithm.
In the SERL framework, it is built on RLPD \cite{ball2023efficient}. 
Throughout the following sections, any reference to DRQ will always imply its use in combination with RLPD.
With this approach, we are able to find an optimal policy for a POMDP $(\mathcal{O, A} , p, r, \gamma)$, using either RGB images, depth images or a voxel grid as additional observation inputs.
These high dimensional observations spaces and their corresponding encoder will be further described in the upcoming \cref{sec:methods}.

\begin{figure}[t!]
  \centering
  \includegraphics[width=0.95\linewidth]{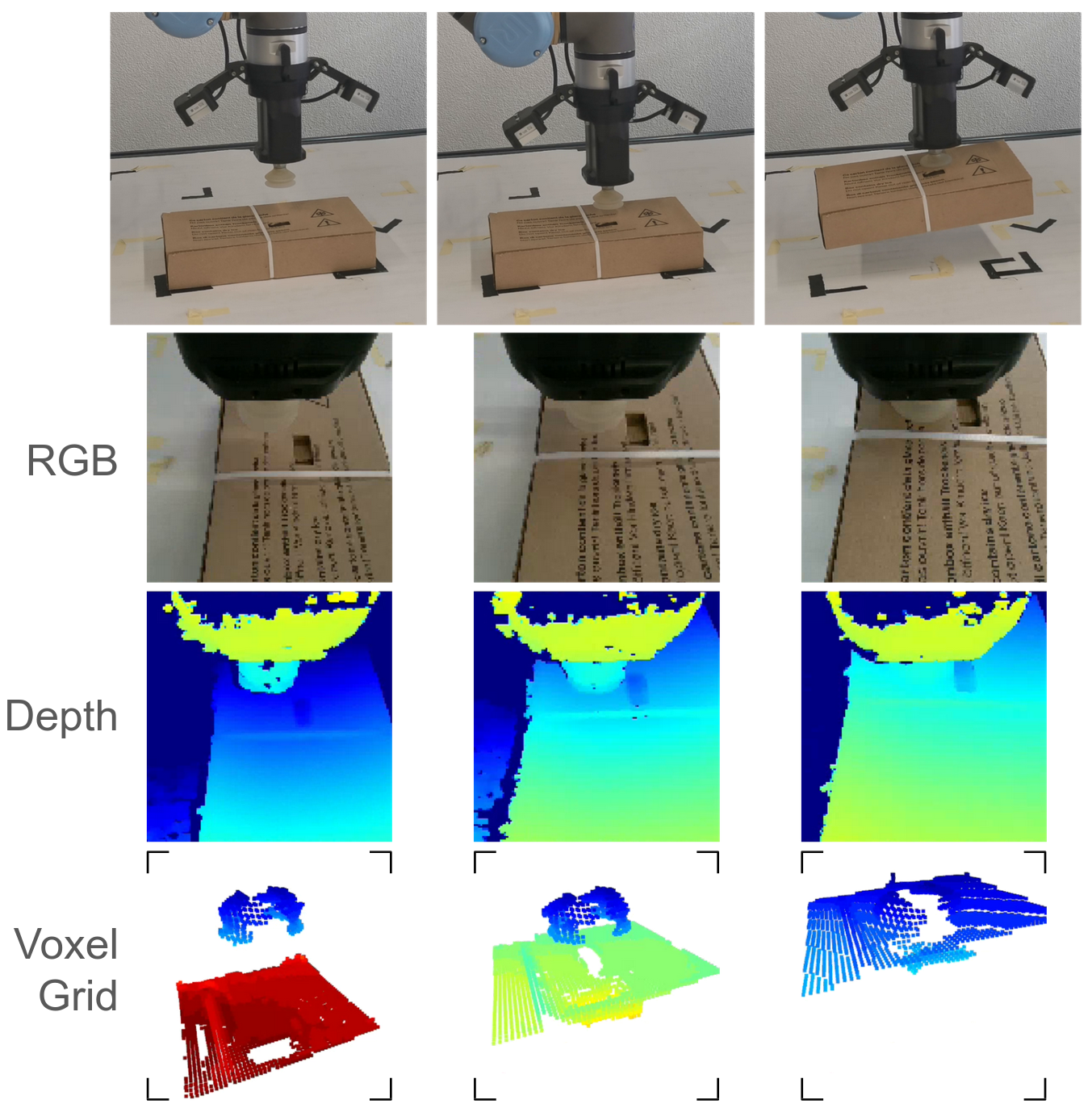}
  \caption{Observation space examples at 3 time-steps during a trajectory.}
  \label{graphic:trajectory timeline}
\end{figure}

\section{Methodology}
\label{sec:methods}

In the following, we describe the environment and the problem setup on which we evaluate the different visual and spatial inputs. 
Also, we describe the encoders that were used for their corresponding high dimensional inputs. 
Lastly, we additionally investigated how leveraging observation space symmetries can further enhance the performance of our policies.

\subsection{UR5 Controller}
Similar to \cite{luo2024serl}, we use an impedance controller for the UR5e Robot Arm \cite{ur5robot}, also employing a two-layered control hierarchy. 
The first layer is the RL policy, running at a lower frequency (\SI{10}{\hertz}), providing target poses for the second, low-level impedance controller running at \SI{100}{\hertz}. 
In our case, the control objective for this controller is
\begin{equation}
F = k_p \cdot e + k_d \cdot \dot{e}
\end{equation}
where the error $e$ is defined as $e = p - p_\text{ref}$, with $p$ being the measured pose, and $p_\text{ref}$ being the target pose. 
The forces are in task space and are sent to the robot arm via the Universal Robot API, which converts them to joint space torques.
Following \cite{luo2024serl}, we bound $e$ such that $|e| \leq \Delta$, avoiding a hard collision or damage when the arm is in contact.

\subsection{Environment}
\label{sec:environment}
The base environment consists of a 7-dimensional action space and a 27-dimensional observation space. 
In the following, the observation space is extended by multiple different high-dimensional representations, such as images, depth images or a voxel grid depending on which policy is used (discussed in detail in \cref{subsec:visEnc}).

The action space consists of the position-delta $\Delta p \in \mathbb{R}^3$, rotation-delta $\Delta \phi \in \mathbb{R}^3$ and if the suction cup should activate, release or stay inactive. 
The observation space consists of (1) the 6D relative pose of the end-effector; (2) the 6D velocity of the end-effector; (3) 3D forces and 3D torques acting upon the tool tip; (4) 2D gripping information consisting of suction cup pressure and grasp information; and (5) the previously executed action.

We followed the environment layout used in SERL, but use Modified Rodrigues Parameters \cite{crassidis1996mrp} in place of Euler angles for describing the orientation in both action and observation space.
This enables rotations to be independent of rotation axes and makes them singularity free (see \cite{geist2024learning} for more analysis on different representations of rotations for learning).
In addition, we extend the work on the relative observation and action frame by \cite{luo2024serl}, transforming the force and torque acting upon the suction gripper as well. 
The observations and actions therefore always represent the state relative to the starting position. 

\subsubsection*{Rewards}
As mentioned above, we assume that we have access to a possibly inaccurate estimate of the position and orientation of the box we want to pick.
Sources of such inaccuracy are, e.g., obstruction of the camera view by the arm once we are close, or a deformation of the box.
The goal is reached if the box is picked and the relative pose is \SI{1}{\centi\meter} above the starting pose, resulting in a reward $R_\text{goal}=100$. 
Furthermore, the agent is linearly penalized for (1) $R_\text{pose}$: deviating too far from the starting pose (both position and orientation); (2) $R_\text{action}$: large actions and action-differences using the euclidean norm; (3) $R_\text{step}$ for each step; and (4) $R_\text{suction}$ for unnecessarily activating the suction cup, while giving a minor reward for a successful grip per step.
\begin{equation}
R(s) = R_\text{goal} - R_\text{step} - R_\text{pose} - R_\text{action} + R_\text{suction}
\end{equation}
Notice that $R_\text{suction}$ can be positive or negative, while the other terms are always larger than zero. 

\subsection{Visual Encoder}
\label{subsec:visEnc}
\textbf{RGB images.} 
The image observations from the wrist cameras of the robot extend our state observation and are passed through a visual backbone before connecting it with the policy MLP.
Compared to SERL, which uses a pre-trained ResNet10, we use a pre-trained ResNet18 \cite{paszke2019pytorch} as this performed better in our case (see \cref{sec:ablations} for more details).

We utilize the pre-trained weights up to the classification layer and keep them fixed during training. Where in the original ResNet \cite{he2016deep} implementation, they average over the spatial dimensions, other pooling methods which can capture the spatial relations are used instead.
While SERL used \textit{Spatial Learned Embeddings}, it did not perform as well in our experiments, and we therefore used \textit{Spatial Softmax} \cite{levine2016ssoftmax}.

During development, we extensively analyzed different pooling methods as well as backbone configurations.
Additionally, we tested using the images of 2 wrist-mounted cameras. 
The training with 2 images simultaneously did not converge, even with more sparse representations like gray-scale images.
Therefore, we trained the image based policy on one image only.
This limitation could be due to hardware constraints, as our GPU is less capable than the one used in SERL, exact specifications are given in \cref{sec:experiments}.

\textbf{Depth images.}
Using depth images is a natural extension to using RGB images only when hoping to capture more spatial information for the learned policy.
We fix the maximum distance of the depth image to \SI{20}{\centi\meter}, since we only want to be informed of the local surroundings of the suction gripper. 

While there is some work on using a ResNet \cite{he2016resnet} with depth images only \cite{katrolia2021depthresnet}, most methods focus on using \mbox{RGB-D} data \cite{wang2018depth, gupta2014learning}.
Thus, we used a simple encoder based on 2D convolutional layers, coupled with the same pooling method as mentioned above.
This is due to the scarce amount of real-life data available to train the encoder, in parallel to training the optimal policy.
The complexity is vastly reduced, resulting in $\sim 1\%$ of the trainable weights compared to a standard ResNet18 model.

\subsection{Spatial Encoder}
\label{sec:spactial_encoder}
Another possible representation for spatial information is a voxel grid, a 3D data structure that represents a volume in space by dividing it into small discrete cubes called voxels. 
Each voxel in the grid corresponds to a specific region of the space, where occupied voxels represent that there is an object in that region of space.

In order to generate the voxel grid representation, we first, combine the point clouds from both cameras using fine-tuned parameters which capture the spatial relation of the cameras, utilizing Open3D's \cite{Zhou2018open3d} multiway registration in the process.
Afterwards, the voxel grid is computed from the combined point cloud: a voxel is occupied if at least one point is within a single voxel.
Since the cameras are fixed on the end effector, with this approach, the voxel grid is always relative to the suction gripper, effectively normalizing the point cloud in end-effector frames \cite{liu2022relw_frame}. 
We limit the voxel grid to the local surrounding of the suction gripper, which is straightforward with this representation. 

\subsubsection*{Architecture}
We chose a 3D convolutional architecture inspired by VoxNet \cite{maturana2015voxnet} over the more recent PointNet \cite{qi2017pointnet}. This choice was primarily driven by the absence of existing PointNet implementations in JAX \cite{jax2018github}, making the implementation of 3D convolutions significantly more straightforward.
VoxNet's simplicity further streamlined our development process, providing a balance between ease of integration and sufficient performance for our needs. This combination of factors made VoxNet the more suitable option for our project.

In the simple spatial encoder, we do not use pretrained weights and train the encoder while updating the policy. 
Further tests revealed that using pre-trained VoxNet weights (from \cite{GitHub2018VoxNet}) can lead to performance increases, since simultaneously learning the policy and encoder weights is not optimal, respectively fully training the VoxNet backbone requires more time and data than is possible to obtain in the real world-RL setting.

Following \cite{yarats2021imageDRQ}, we extended the data augmentation technique to the voxel grid, allowing a random shift in 3 dimensions.
In our implementation of this method, the voxel grid is padded on each side by 3 empty voxels, and then randomly cropped back to the original size.
Future work may include using additional 3D augmentation methods, such as occlusion or noise injection.

\begin{figure}[tb]
  \centering
  \includegraphics[width=0.8\linewidth]{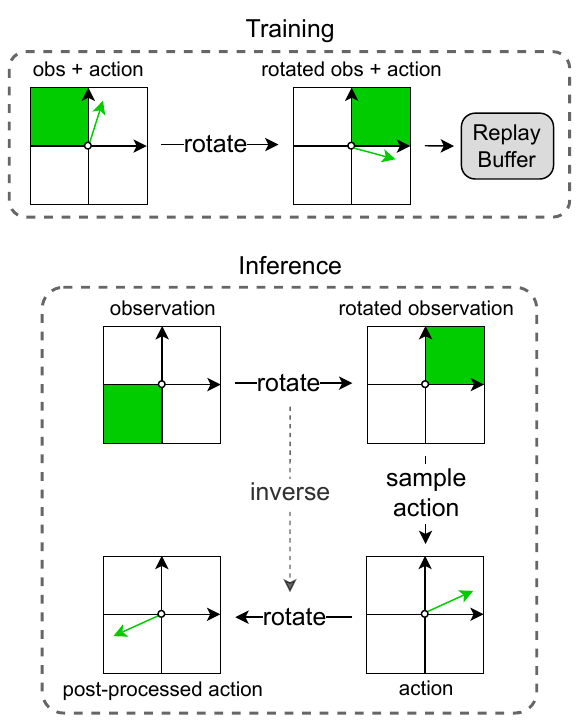}
  \caption{Observation (obs) and action rotation behavior during training and inference.}
  \label{graphics:symmetries}
\end{figure}

\subsection{Observation Space Symmetries}
\label{sec:obs_space_symmetries}
As both the reward, as well as the orientation, mentioned in \cref{sec:environment}, are rotation-invariant, a trajectory rotated by 90 degrees around the $z$-axis is effectively equivalent.
Tests on using the symmetry for augmentation purposes, essentially quadrupling the sampled data, did not result in performance increases, increasing the training times of policies significantly.

While there is some work on exploiting symmetries in RL \cite{gupta2022exploiting, kamal2008reinforcement} to further improve sample efficiency, we did not employ these methods directly.
Instead, noticing the 4-fold rotation symmetry of the observation-, action- and voxel grid space around the $z$-axis, we transformed every observation in the real world into the first and fifth octant, respectively.
As an example for the use symmetries: the position deviation is streamlined, since $X$ and $Y$ deviations will always be positive when exploiting the symmetries, and a correction of it will always be a negative action.
As a result, the policy is not forced to learn each quadrant in the $XY$-plane separately, simplifying the problem statement.
\cref{graphics:symmetries} provides information on the training and inference usage of the method, and in \cref{tab:policy_infos} it is denoted as \textit{Uses Symmetries}.

\section{Experiments}
\label{sec:experiments}

\def\arraystretch{1.2}
\begin{table*}[!ht]
    \centering
    \caption{Policy specifications and training times.}
    \begin{tabular}{l l c c c c }
        \specialrule{.1em}{.05em}{.05em} 
        ID & \textbf{Policy} & \textbf{Backbone} & \textbf{Uses Symmetries} &  \textbf{Additional Observations} &  \textbf{Training Time}\\
        \specialrule{.1em}{.05em}{.05em} 
        (1) & Behavior Tree & - & \xmark & -  & - \\
        \arrayrulecolor{mygray}
        \hline
        (2) & Behavioral Cloning & - & \xmark & - &  ~1 minute \\
        \hline
        (3) & Soft Actor Critic & - & \xmark & - & 17 minutes \\
        \hline
        (4) & Data-regularized Q & pre-trained ResNet18 & \xmark & 1x wrist RGB Image & 50 minutes \\
        \hline
        (5) & Data-regularized Q & Conv2D based Encoder & \xmark & 2x wrist Depth Images & 58 minutes \\
        \hline
        (6) & \multirow{2}{*}{Data-regularized Q} & \multirow{2}{*}{Conv3D based Encoder} & \xmark & \multirow{2}{*}{Voxel Grid} & 46 minutes \\
        (7) & & & \cmark & & 73 minutes \\
        \hline
        (8) & \multirow{2}{*}{Data-regularized Q} & \multirow{2}{*}{pre-trained VoxNet} & \xmark & \multirow{2}{*}{Voxel Grid} & 65 minutes \\
        (9) & & & \cmark & & 85 minutes \\
        \hline
    \end{tabular}
    \label{tab:policy_infos}
\end{table*}

\def\arraystretch{1.2}
\begin{table*}[!ht]
\sisetup{detect-weight=true}
    \centering
    \caption{Policy evaluation. See \cref{tab:policy_infos} for details of the policies. Bolded numbers indicate the best result. Pre-trained voxel grid encoders and symmetry usage is denoted with \textit{pretr.} and \textit{sym.}, respectively. }
	 \begin{tabular}{l l p{1pt} l | S | S s | S s || S | l s | S s}
		\toprule
        \multirow{2}{*}{\small{ID}} & \multirow{2}{*}{\small{Policy}} & \multirow{2}{*}{\rotatebox{90}{\footnotesize{pretr.}}} & \multirow{2}{*}{\rotatebox{90}{\footnotesize{sym.}}} & \multicolumn{5}{c||}{\small{seen boxes}} & \multicolumn{5}{c}{{\small{unseen boxes}}} \\
		& & & & {Success [\%]} & \multicolumn{2}{l|}{Reward} & \multicolumn{2}{l||}{Time [s]} & {Success [\%]} & \multicolumn{2}{l|}{Reward} & \multicolumn{2}{l}{Time [s]} \\
		\midrule
		(1) & BT & & & 86.7 & 47.6 & $\pm29.5$ & 4.8 & $\pm1.7$ & 86.7 & 47.6 & $\pm32.8$ & 5.32 & $\pm1.4$ \\
		(2) & BC & & & 80 & -3.5 & $\pm20.8$ & 5.03 & $\pm1$ & 60 & -24.6 & $\pm5.5$ & 7.22 & $\pm1$ \\
		(3) & SAC & & & 73.3 & 36.9 & $\pm18.5$ & 5.59 & $\pm2.1$ & 50 & -21.2 & $\pm16.6$ & 7.48 & $\pm2$ \\
		(4) & DRQ RGB & & & 96.7 & 58.5 & $\pm57$ & 4.08 & $\pm1.7$ & 80 & -22.3 & $\pm28.2$ & 5.28 & $\pm1.1$ \\
		(5) & DRQ Depth & & & \bfseries 100 & 83.8 & $\pm14.1$ & 3.01 & $\pm0.7$ & 86.7 & -25.8 & $\pm23.1$ & 5.32 & $\pm0.9$ \\
        (6) & DRQ Voxel & & & \bfseries 100 & 86.7 & $\pm12.3$ & 2.95 & $\pm1.1$ & 66.7 & 11.1 & $\pm11.9$ & 5.74 & $\pm0.8$ \\
		(7) & DRQ Voxel & & \cmark & 93.3 & 74.4 & $\pm12.2$ & 3.3 & $\pm0.8$ & 83.3 & 19 & $\pm27.3$ & 4.95 & $\pm1.1$ \\
		(8) & DRQ Voxel & \cmark & & \bfseries 100 & \bfseries 90.8 & $\pm4.2$ & \bfseries 2.74 & $\pm0.5$ & 93.3 & 55.4 & $\pm10.1$ & \bfseries 3.88 & $\pm0.5$ \\
		(9) & DRQ Voxel & \cmark & \cmark & 96.7 & 83.9 & $\pm11.6$ & 2.84 & $\pm0.7$ & \bfseries 96.7 & \bfseries 65.3 & $\pm30.7$ & 4.34 & $\pm1.6$ \\
		\bottomrule
		\end{tabular}
  \vspace{-3mm}
        \label{tab:results}
\end{table*}

For the experiments, we use a 6-DoF UR5 Robot Arm \cite{ur5robot} with the Robotiq EPick vacuum gripper \cite{robotiqEpick}. 
On the side of the end-effector, 2 Intel RealSense D405 \cite{keselman2017realsense} cameras are placed (as can be seen in \cref{fig:intro}) for its vision capabilities.
Each environment interaction is limited to 100 steps, equivalent to \SI{10}{\second}, after which the trajectory is truncated.

As baselines, we use a simple behavior tree (\cref{sec:BT}) alongside behavioral cloning (\cref{sec:BC}). 
The behavioral cloning model is trained using 20 expert demonstrations, the same that are used for the other policies.

Observations include camera images, depth images, or voxel grid representations, depending on the policy used, as well as end-effector pose, twist, force, torque, gripping information and the last action performed. 
The image size for RGB and depth images are set to 128x128x3 and 128x128x1, respectively.
We further cap the maximum distance for the depth image to be \SI{20}{\centi\meter}, since we only want to detect local details.
For the voxel grid, the resolution is set to 50x50x40 voxels, with a voxel size of 2x2x2\,\si{\cubic\milli\meter} resulting in a coverage of 10x10x8\,\si{\cubic\centi\meter} around the suction cup.
While this limits the information that is available to the policy, this is in accordance to our assumption of dealing with last-inch manipulation: we only need local information to decide where exactly to grasp, respectively to avoid locations where we can not grasp.

\begin{figure}[t]
  \centering
  \includegraphics[width=0.485\linewidth]{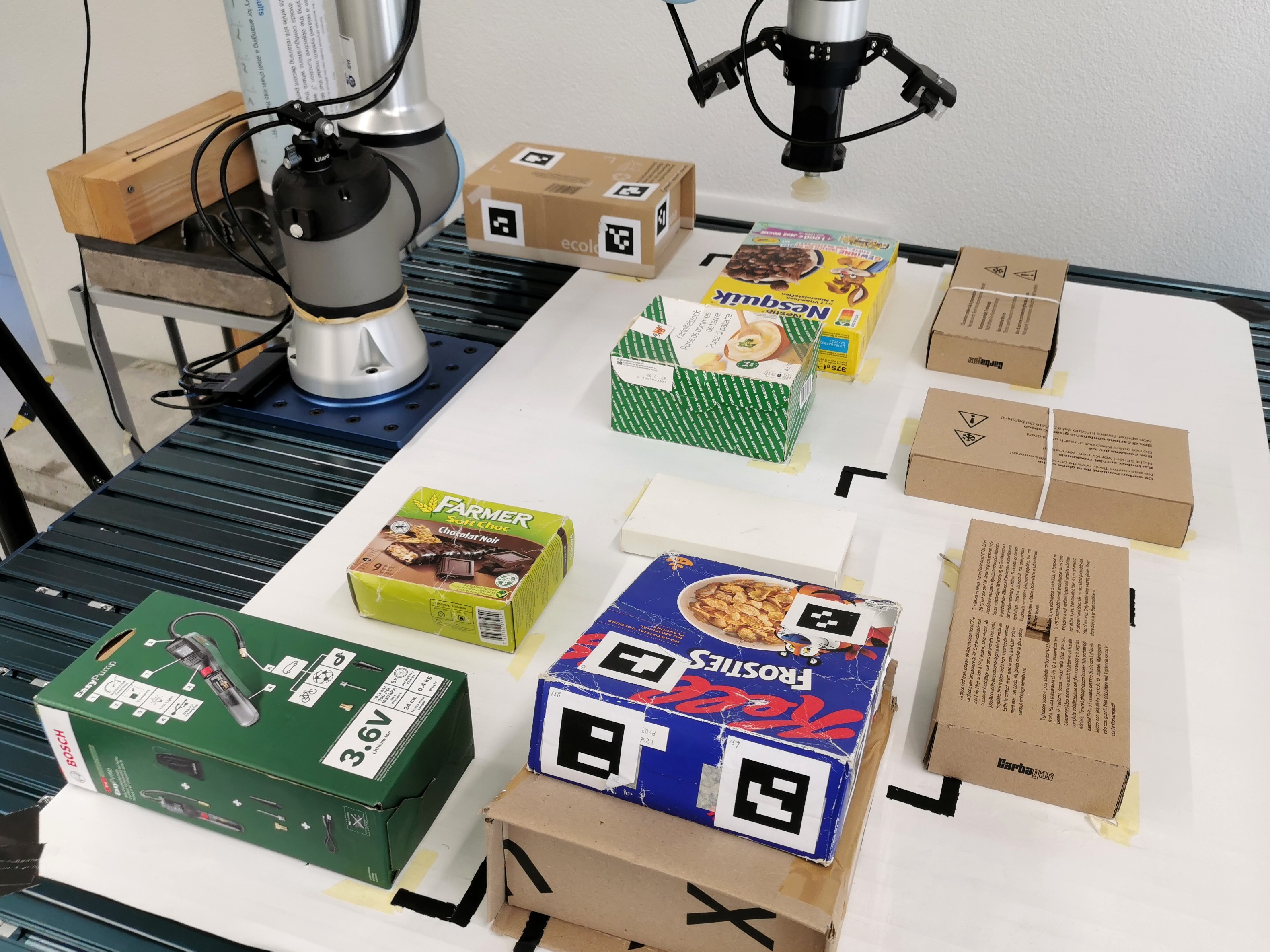}%
  \hspace{0.1mm}
  \includegraphics[width=0.485\linewidth]{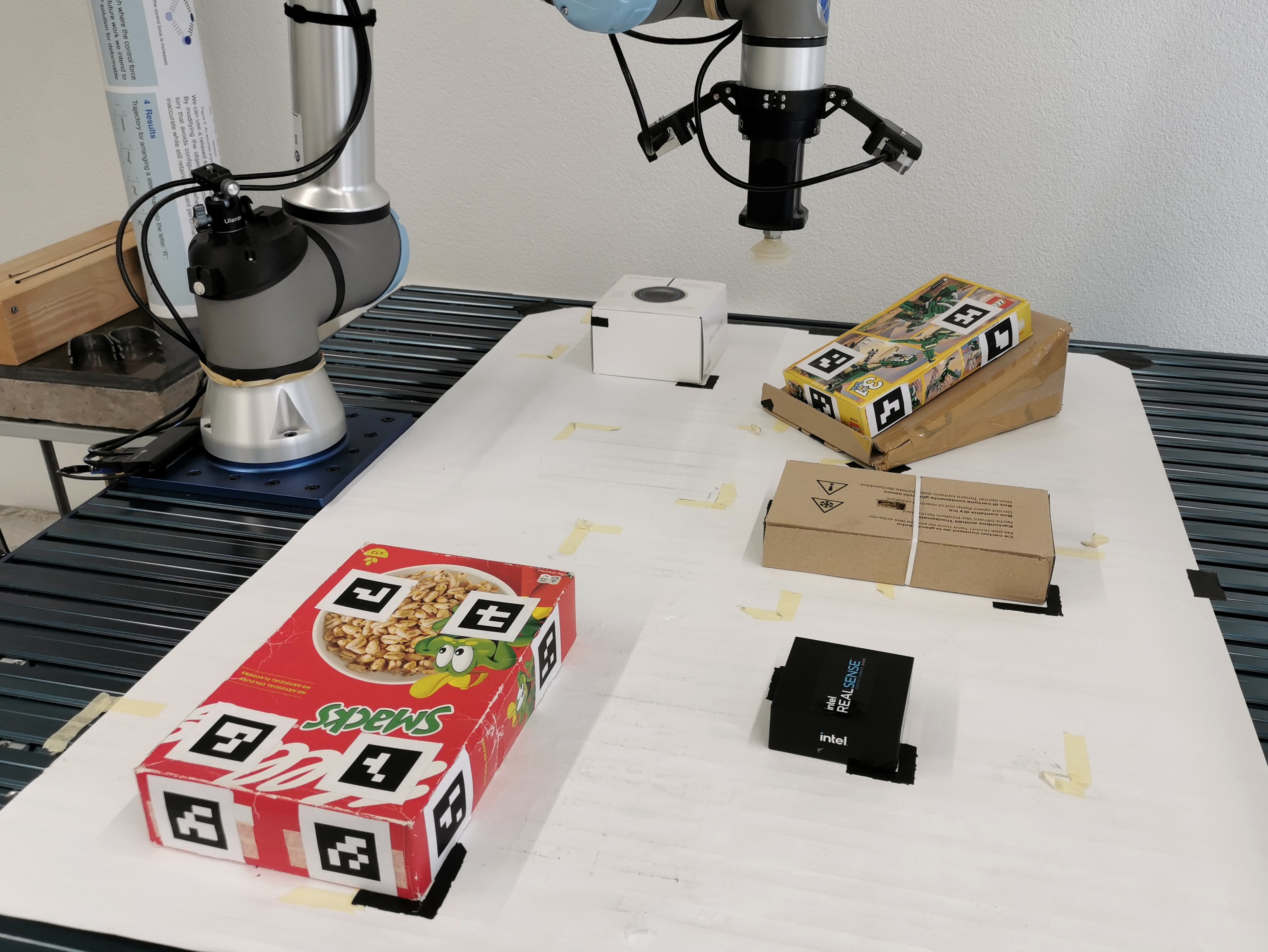}
  \caption{Experiment setup for training (left) and testing (right).}
  \label{graphic:setup}
\end{figure}

\textbf{Policy Training} was done using an environment consisting of 10 different boxes, each at a distinct position (\cref{graphic:setup}) and orientation.
We place them at these predetermined positions in order not to rely on a vision system for determining their location in space, and thus to avoid a further variable in the training\footnote{On the website, we apply the policy to an unstructured setting where we use the policy to clean up a pile of boxes as a proof of concept, relying on an external perception system to obtain an estimate of the poses of the boxes.}.
Note that there is some domain randomization, as the boxes are lifted during training, and are never placed perfectly back at the same location as they were originally. 
Additionally, the end-effector's starting position and orientation is slightly randomized to account for estimation errors that occur in a real system.
The boxes differ in color, size, rigidity, surface texture, and obstructions, which include two boxes with zip-ties, a deformed, concave box, and one box with a hole in it.
Each RL-based policy was initialized with 20 successfully teleoperated demonstrations. 
These demonstrations are fed into the replay buffer at the start of training, providing the agents with multiple successful environment interactions. 
More information on each investigated policy can be found in \cref{tab:policy_infos}.
All training was performed on a single Nvidia RTX 3080 GPU and ran until convergence, with no policy taking longer than $90$ minutes to complete.

The \textbf{Evaluation Setup } consists of two scenes (\cref{graphic:setup}):
(1) a setup using the same boxes as the policy was trained on for a total of 30 trials per policy, and  
(2) a setup with 5 unseen boxes to test generalization, also for 30 trials per policy.
These unseen boxes have different colors and different sizes and orientations (i.e., one box is tilted).
Further, one box has a zip tie and hole in it which needs to be avoided for a successful grasp, and for one box, we initialize the end effector with a substantial angle relative to the box, simulating a wrong state estimate.
For a fair comparison, the initial end effector pose is seeded and sampled uniformly in both setups, in order to guarantee the same initial conditions for each evaluated policy.

The \textbf{evaluation metrics} we used are  
(1) The success rate, which is 1 for a success and 0 for a failure, with no partial credits; (2) the cumulative reward of the trajectory executed given by the environment, as described in \cref{sec:environment}; and (3) the time it takes for a trajectory to reach the goal state or time out.

\subsection{Results.}
\cref{tab:results} presents the detailed results for the box picking task in both setups, with one standard deviation for both reward and time.
On the seen boxes, where the policies were trained on, policies that depend on point cloud data (5)-(9) plus ResNet18 as backbone (4) perform well on the evaluation, having $90\%+$ success rate. 
We can also see that the time required for the box picking decreases for point cloud based policies (5)-(9), staying around $3$s per picking task versus $4$s for the image based approach (4) and $5$s for the baseline policies (1)-(3). 

In the evaluation on the unseen boxes, we see that the task is in fact similar to the first (i.e., the evaluation on seen boxes), with BT (1) showing a very similar performance. 
We notice that BC, SAC and image based DRQ (2) - (4) did not perform well, with a significant success rate decrease relative to the evaluation on the seen boxes. 
Approaches that use point cloud observations (5) - (9) perform better overall, excluding policy (6), not handling the new boxes as well.
The biggest shift can be seen in the cumulative reward: While on seen boxes, policies (3) - (5) perform similarly, on unseen boxes the voxel based approaches (6) - (9) clearly surpass the former.
That change is also noticeable in the mean time taken, where the policies using a pre-trained VoxNet as a backbone (8) and (9) have the upper hand in the unseen boxes evaluation.

\subsection{Discussion.}
While BT (1) and BC (2) set the baseline with a success rate of $87\%$ and $80\%$, we can see that SAC (3) trained only on proprioceptive states did not increase performance on this task, neither on the trained boxes nor the unseen ones, therefore suggesting that there is a need for some visual or spatial information.

Although image based DRQ (4) has a high success rate of $97\%$ on the training boxes, it does not generalize well, achieving only $80\%$ on the unseen ones, lower than BT (1).
We can observe that the image policy (4) learned how to pick the known boxes well, but without learning to generalize on other boxes.
One possibility for the failure to generalize might be that the boxes are over-reliant on the color of the boxes.
We did briefly examine a policy similar to (4) but with grayscale images, which did however not change the results by much.

On the other hand, for both depth image and voxel based approaches (5) - (9), the success rate does not decline drastically, with the exception being policy (6). 
The tests revealed a significant relationship between backbone choice and generalization performance, with voxel based policies outperforming former approaches.
Comparing (6) and (8) suggests that the problems of (6) might originate from an undertrained backbone.

Comparing the policies (6) and (7) with their pre-trained counterpart (8) and (9), we clearly see that while the training time increases, the success rate on unseen boxes is higher with the latter $(93\%$ vs $67\%$ and $97\%$ vs $83\%)$, indicating that using pre-trained weights positively affects policy generalization.

Using the observation space symmetry (\cref{sec:obs_space_symmetries}), the tests revealed that using the method further increases both success rate and reward for policies (8) and (9) when evaluating on unseen boxes ($93\%$ vs  $83\%$ and $97\%$ vs $93\%$).

\subsection{Ablations}
\label{sec:ablations}
The ablations policies are defined in \cref{tab:ablation_spec}, while \cref{tab:no_proprio} shows the evaluation results in comparison to the respective baselines.

\def\arraystretch{1.2}
\begin{table}[h!]
    \centering
    \caption{Ablation policy specifications.}
	\begin{tabular}{c | c | c }
    \toprule
    ID & Baseline & Key Changes \\ 
    \midrule
    (10) & (4) & Pretrained ResNet10 instead of ResNet18.\\ 
    (11) & (9) & Limited proprioception (only gripper information). \\
    (12) & (9) & Enabled temporal ensembling.  \\ 
    (13) & (11) & Enabled temporal ensembling.  \\ 
    \bottomrule
  \end{tabular}
    \label{tab:ablation_spec}
\end{table}

\textbf{ResNet10 Encoder.}
As mentioned in \cref{subsec:visEnc}, using a pre-trained ResNet18 \cite{he2016resnet} as backbone improves the agents performance compared to the ResNet10 used originally in SERL. 
\cref{tab:no_proprio} shows that the performance of policy (10) decreases considerably.
We suspect that this is mainly due to the objects in different SERL tasks having distinct, homogeneous features (e.g., PCB board, red tomato) compared to our heterogeneous environment.

\def\arraystretch{1.2}
\begin{table}[h!]
    \centering
    \caption{Ablation evaluations. Values in parentheses are the delta between the ablated policies and their respective baseline.}
	\begin{tabular}{l | l | c l l}
    \toprule
    Policy & Experiment & Success [\%] & Reward & Time [s] \\ 
    \midrule
    (10) & seen boxes & 66.7 (\textcolor{red}{-30}) & 14.7 (\textcolor{red}{-43.8}) & 5.24 (\textcolor{red}{+1.16}) \\
    
    (10) & unseen boxes & 40 (\textcolor{red}{-40}) & -54.7 (\textcolor{red}{-32.4}) & 7.62 (\textcolor{red}{+2.34}) \\
    \hline
    (11) & seen boxes & 93.3 (\textcolor{red}{-3.3}) & 82.2 (\textcolor{red}{-1.7}) & 2.86 (\textcolor{red}{+0.02}) \\
    
    (11) & unseen boxes & 100 (\textcolor{ForestGreen}{+3.3}) & 74.8 (\textcolor{ForestGreen}{+9.5}) & 3.91 (\textcolor{ForestGreen}{-0.43}) \\
    \bottomrule
  \end{tabular}
    \label{tab:no_proprio}
\end{table}

\textbf{Voxel Grid only.}
We tested training a policy only relying on voxel grid observations and 2D gripper information (decreasing the agents state dimensions from 27 to 2), and the policy performed surprisingly well.
The results of policy (11) can be seen in \cref{tab:no_proprio}.

The approach scored $100\%$ on the unseen evaluation success, even surpassing the original policy (9) with the same specifications plus full state information.
These results show a significant relationship between voxel grid observation and the resulting action, demonstrating that the spatial encoder is capable of perceiving its local environment, and possibly suggesting that many of the proprioceptive states are redundant with the spatial observations.

\textbf{Temporal ensembling.}
While large deviations from previous actions are penalized in the reward function, the resulting action sequences can still be jittery, resulting in high force oscillations, decreasing the lifetime of the robot arms motors.
To increase the smoothness of the trajectory, temporal ensembling \cite{zhao2023aloha} can be used, in our case given by:
\begin{equation}
\label{equation:temporal ensembling}
a_t = [a^s_{t}, a^s_{t-1}, a^s_{t-2}, a^s_{t-3}] \times [0.5, 0.3, 0.2, 0.1]^T
\end{equation}
where $a^s_t$ is the action sampled from the policy at time-step $t$ and $a_t$ is the action which will be executed.
We re-evaluate policies (9) and (10) on the unseen boxes with temporal ensembling activated and obtain the results in \cref{tab:temporal_ens}.

\def\arraystretch{1.2}
\begin{table}[h!]
    \centering
    \caption{Evaluation with temporal ensembling on the unseen boxes. Values in parentheses are the delta between the ablated policies and their respective baseline.}
	\begin{tabular}{l | l l l l}
    \toprule
    Policy & Success [\%] & Reward & Time [s] & Smoothness $\bar{d}$\\ 
    \midrule
    (12) & 96.7 (+0.0) & 64 (\textcolor{red}{-1.3}) & 4.36 (\textcolor{red}{+0.02}) & {1.19} (\textcolor{ForestGreen}{-1.27}) \\
    \hline
    (13) & 96.7 (\textcolor{red}{-3.3}) & 67.3 (\textcolor{red}{-7.5}) & 4.03 (\textcolor{red}{+0.12}) & {1.19} (\textcolor{ForestGreen}{-1.03}) \\
    \bottomrule
  \end{tabular}
    \label{tab:temporal_ens}
\end{table}

Since continuous actions are smooth, we use action difference $d = \sum_{t=1}^{t_{max}} || a_t - a_{t-1} ||_2$ averaged over all trajectories as an indicator.
While it is obvious from equation \cref{equation:temporal ensembling} that the smoothness improves, it is more important to highlight that the new policy performs almost as well, with only minor effects on success rate, reward, and time, even though this modification was not present during training, but only added later at evaluation-time.

\section{Conclusion}
\label{sec:conclusion}

To the best of our knowledge, our approach is the first work to train a reinforcement learning policy for a suction gripper using point cloud inputs with real world data. 
In our work, we conducted a comparison between well-established 2D image representations against 3D point cloud based representations.
Through the real world example of box picking, we showed that 3D voxel grid representations are particularly beneficial for spatial perception, outperforming both the behavior tree baseline, and all image based policies.

In future work, we want to expand this work to more tasks including different manipulation tasks such as dual-arm manipulation, or cooperative handovers and compare the VoxNet inspired architecture to a PointNet \cite{qi2017pointnet} or Point transformer \cite{zhao2021pointtransformer} based encoder. 

Focusing more on the task at hand, it might be interesting to apply the real-world-RL approach that we took in this work to, e.g., \cite{ankile2024imitationrefinementresidual} which learns residual actions from a BC policy, instead of learning the complete policy from scratch in the real world.

\section*{Acknowledgment}

The authors thank Miguel Angel Zamora Mora, Simon Huber and Moritz Geilinger for their valuable input.

\bibliography{bibliography/references}

\end{document}